\documentclass[10pt,twocolumn,letterpaper]{article}

\usepackage[pagenumbers]{cvpr} 

\definecolor{cvprblue}{rgb}{0.21,0.49,0.74}
\usepackage[pagebackref,breaklinks,colorlinks,allcolors=cvprblue]{hyperref}

\usepackage{url}
\usepackage{multirow} 
\usepackage{graphicx}
\usepackage[export]{adjustbox}
\usepackage{float}
\usepackage{epsfig}
\usepackage{graphicx}
\usepackage{amsmath}
\usepackage{amssymb} 
\usepackage{caption}
\usepackage{xparse} 
\usepackage{wrapfig} 
\usepackage{tabularx}
\usepackage{subcaption}
\usepackage{fancyhdr}
\usepackage{color}
\usepackage{wrapfig} 
\usepackage{framed}
\usepackage{soul}
\usepackage{xspace}
\usepackage{booktabs} 
\usepackage{makecell}
\usepackage{xcolor}
\usepackage{vcell}
\usepackage{colortbl}
\usepackage[nodisplayskipstretch]{setspace}
\usepackage{seqsplit} 
\usepackage{array}
\usepackage{CJKutf8} 
\usepackage[normalem]{ulem}
\usepackage{enumitem}
\usepackage{etoc}
\usepackage{tcolorbox}
\usepackage{marvosym}
\newcommand{\header}[1]{\text{#1}}

\definecolor{backred}{RGB}{255, 190, 190}
\definecolor{backblue}{RGB}{210, 230, 250}
\definecolor{myblue}{RGB}{6, 174, 226}

\definecolor{darkgreen}{rgb}{0.0,0.5,0.0}

\definecolor{shadecolor}{RGB}{237,237,237}

\definecolor{uclablue}{rgb}{0.15, 0.45, 0.68}
\definecolor{Gray}{gray}{0.93}
\definecolor{uclagold}{rgb}{1.0, 0.7, 0.0}
\definecolor{airforceblue}{rgb}{0.36, 0.54, 0.66}
\definecolor{rosegold}{rgb}{0.72, 0.43, 0.47}
\definecolor{pastelbrown}{rgb}{0.51, 0.41, 0.33}
\definecolor{isabelline}{rgb}{0.96, 0.94, 0.93}
\definecolor{macaroniandcheese}{rgb}{0.98, 0.89, 0.83}
\definecolor{wildblueyonder}{rgb}{0.85, 0.89, 0.95}
\definecolor{mediumtaupe}{rgb}{0.4, 0.3, 0.28}
\definecolor{bluegray}{rgb}{0.4, 0.6, 0.8}
\definecolor{celestialblue}{rgb}{0.29, 0.59, 0.82}
\definecolor{darkorange}{rgb}{1.0, 0.55, 0.0}
\definecolor{cadmiumred}{rgb}{0.89, 0.0, 0.13}
\definecolor{magnolia}{rgb}{0.97, 0.96, 1.0}
\definecolor{pastelblue}{rgb}{0.68, 0.78, 0.81}
\definecolor{persiangreen}{rgb}{0.0, 0.65, 0.58}
\definecolor{steelblue}{rgb}{0.27, 0.51, 0.71}
\definecolor{bluebell}{rgb}{0.64, 0.64, 0.82}
\definecolor{dimgray}{rgb}{0.41, 0.41, 0.41}
\definecolor{splashedwhite}{rgb}{1.0, 0.99, 1.0}
\definecolor{lavendergray}{rgb}{0.77, 0.76, 0.82}
\definecolor{lightgray}{rgb}{0.83, 0.83, 0.83}
\definecolor{lavendermist}{rgb}{0.9, 0.9, 0.98}
\definecolor{lightgreen}{HTML}{f8fcf4}
\definecolor{lightblue}{HTML}{dfebf7}
\definecolor{zeroshot}{rgb}{0.9, 0.9, 0.9}
\definecolor{fourshot}{rgb}{0.8, 0.9, 0.8}
\definecolor{eightshot}{rgb}{0.8, 0.8, 0.9}
\definecolor{sixteenshot}{rgb}{0.9, 0.8, 0.8}
\definecolor{blue-violet}{rgb}{0.54, 0.17, 0.89}
\definecolor{coral}{HTML}{FF7F50}
\definecolor{lightyellow}{RGB}{255, 250, 194}
\definecolor{lightred}{RGB}{250, 189, 164}
\definecolor{lightorange}{RGB}{254, 226, 205}

\newcommand{\model}{\textsc{G$^2$VLM}\xspace} 

\definecolor{cvprblue}{rgb}{0.21,0.49,0.74}
\usepackage[pagebackref,breaklinks,colorlinks,allcolors=cvprblue]{hyperref}

\def\paperID{5274} 
\def\confName{CVPR}
\def\confYear{2026}

\title{\model: Geometry Grounded Vision Language Model with Unified  3D Reconstruction and Spatial Reasoning}

\author{
 Wenbo Hu$^{1,2*}$\quad
 Jingli Lin$^{1,3*}$\quad 
 Yilin Long$^{1,4*}$\quad
 Yunlong Ran$^{1,5}$\quad 
 Lihan Jiang$^{1,6}$\quad \\
 Yifan Wang$^{1,3}$ \quad  
 Chenming Zhu$^{1,7}$\quad
 Runsen Xu$^{1,8}$\quad
 Tai Wang$^{1\dagger}$\quad
 Jiangmiao Pang$^{1\dagger}$\quad
\\
$^1$Shanghai AI Lab \quad
$^2$UCLA \quad
$^3$SJTU\quad
$^4$FDU\quad 
$^5$ZJU\quad  
$^6$USTC\quad  
$^7$HKU\quad  
$^8$CUHK\quad 
\\ 
$^{*}$Equal Contribution \quad
$^{\textsuperscript{$\dagger$}}$Corresponding Author\\
\href{https://gordonhu608.github.io/g2vlm.github.io/}{\adjustbox{valign=c}{\includegraphics[height=1em]{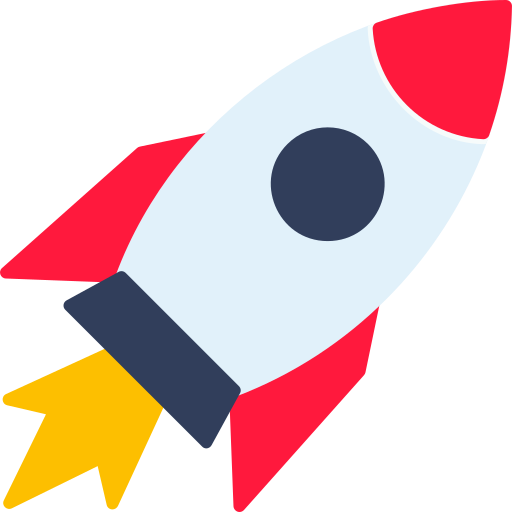}} Project Page} \quad
\href{https://github.com/InternRobotics/G2VLM}{\adjustbox{valign=c}{\includegraphics[height=1em]{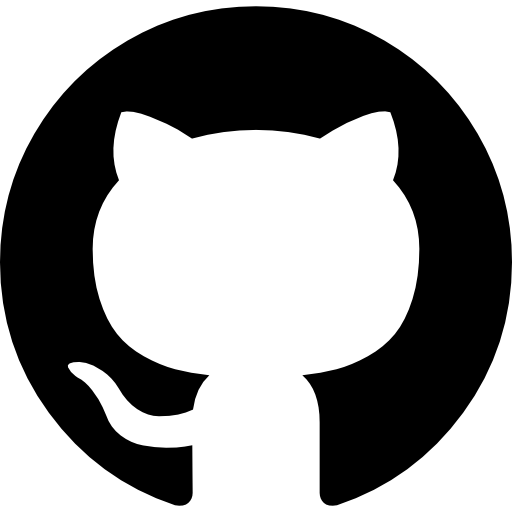}} GitHub} 
}

\begin{document}

\twocolumn[{%
\renewcommand\twocolumn[1][]{#1}%
\maketitle
\begin{center}
    \centering
    \vspace{-1em}
    \captionsetup{type=figure}
   \includegraphics[trim=0cm 17.6cm 1.79cm 0cm, clip, width=\linewidth]{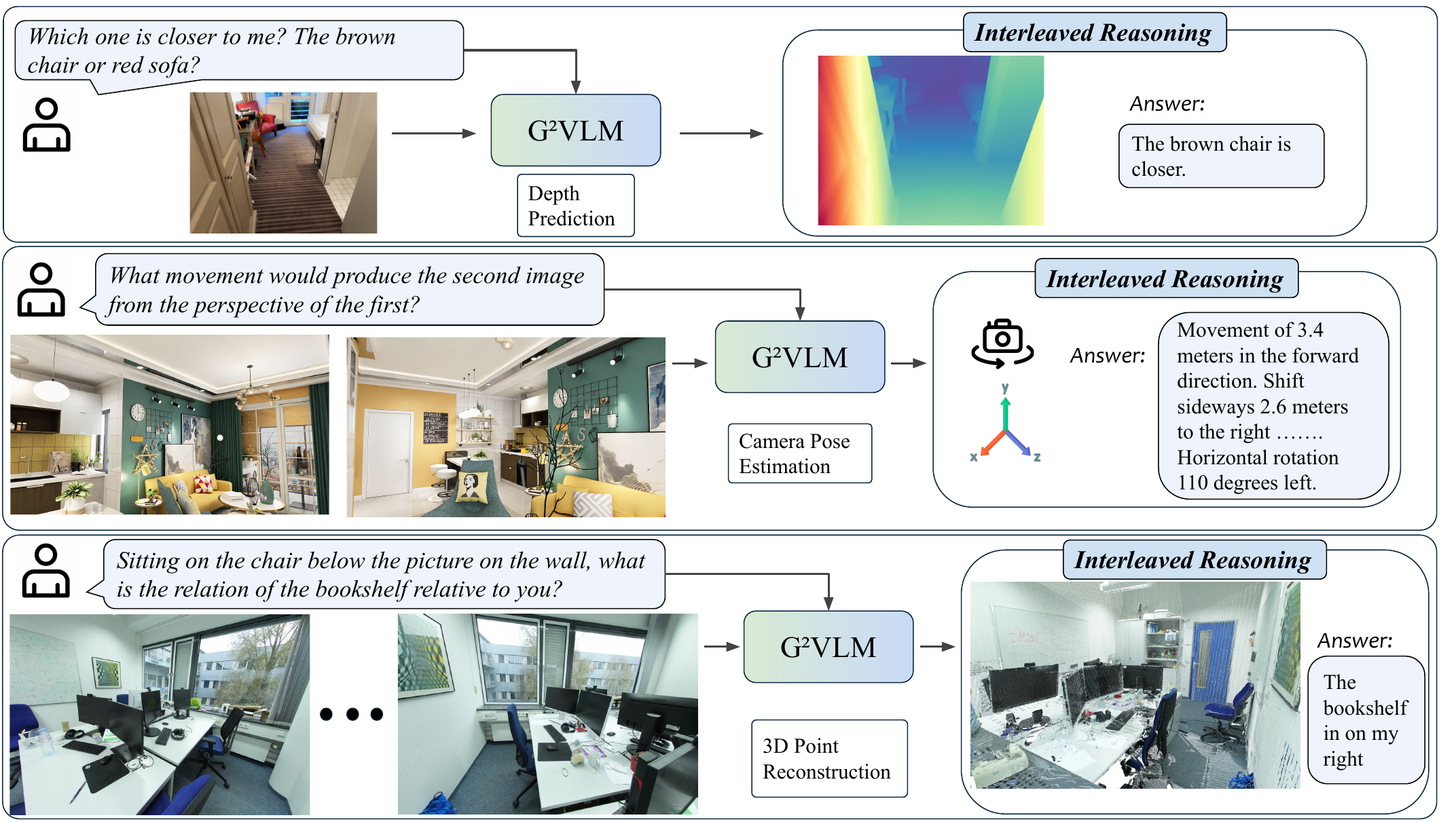}
    \captionof{figure}{We present \model, a geometry grounded vision-language model proficient in both spatial 3D reconstruction and spatial understanding tasks. For spatial reasoning questions, \model can directly predict 3D geometry and employ interleaved reasoning for an answer.}
    \label{fig:teaser}
\end{center}
}]

\vspace{-2mm}
\begin{abstract}

Vision-Language Models (VLMs) still lack robustness in spatial intelligence, demonstrating poor performance on spatial understanding and reasoning tasks. We attribute this gap to the absence of a visual geometry learning process capable of reconstructing 3D space from 2D images. We present \model, a \textbf{g}eometry \textbf{g}rounded \textbf{v}ision-\textbf{l}anguage \textbf{m}odel that bridges two fundamental aspects of spatial intelligence: spatial 3D reconstruction and spatial understanding. \model natively leverages learned 3D visual geometry features to directly predict 3D attributes and enhance spatial reasoning tasks via in-context learning and interleaved reasoning. Our unified design is highly scalable for spatial understanding: it trains on abundant multi-view image and video data, while simultaneously leveraging the benefits of 3D visual priors that are typically only derived from hard-to-collect annotations.
Experimental results demonstrate \model is proficient in both tasks, achieving comparable results to state-of-the-art feed-forward 3D reconstruction models and achieving better or competitive results across spatial understanding and reasoning tasks.
By unifying a semantically strong VLM with low-level 3D vision tasks, we hope \model can serve as a strong baseline for the community and unlock more future applications, such as 3D scene editing. 

\end{abstract}

\section{Introduction}
\vspace{-1mm}

Recent advancements in Vision-Language Models (VLMs) have established them as powerful foundation models across a wide range of tasks~\citep{openai2024gpt4o, zhang2025videollama, bai2025qwen2, wang2025internvl35advancingopensourcemultimodal}.
Yet several studies~\citep{chen2024spatialvlm, yu2025far,du2024embspatial,stogiannidis2025mind} have found that VLMs exhibit a significant gap in robust spatial understanding, a crucial capability for spatial intelligence, robotics, and embodied AI.

We argue that this limitation stems from how current VLMs acquire their physical world knowledge. Their spatial understanding is largely gained implicitly from training on massive, unstructured 2D image-text datasets, relying primarily on language and 2D visual priors. Except incorporating 3D priors as in specific 3D-VLMs~\citep{zhu2024llava, zheng2025video3dllmlearningpositionaware, hu20253dllmmemlongtermspatialtemporalmemory}, general VLMs simply employ feature projection layers and are trained with auto-regressive next-token prediction, which treat inputs like multiple images or video frames as a ``flat'' sequence of 2D data. This approach lacks the explicit learning of visual geometry that can ``lift'' these 2D perceptions into a coherent 3D representation of the world, which is crucial for genuine spatial understanding. 

To overcome this limitation, we propose to integrate visual geometry learning into the VLM. As illustrated in Figure~\ref{fig:teaser}, our method can directly predict 3D outputs and employ them within an interleaved reasoning process to enhance spatial reasoning.
Our approach is inspired by the two-streams hypothesis in human cognition~\citep{Goodale1992Separate, wang2025visuallydescriptivelanguagemodel}, which posits a ``ventral stream'' for object recognition (correlating with multimodal understanding) and a ``dorsal stream'' for spatial location (corresponding to visual geometry learning). 

We introduce \model, a geometry grounded vision-language model that bridges two fundamental aspects of spatial intelligence: spatial 3D reconstruction and spatial understanding. \model natively leverages learned 3D geometry features to enhance spatial reasoning tasks via in-context learning and interleaved reasoning.

\begin{figure}[t]
  \centering
   \includegraphics[trim=0cm 30.5cm 27.2cm 0cm, clip, width=0.7\linewidth]{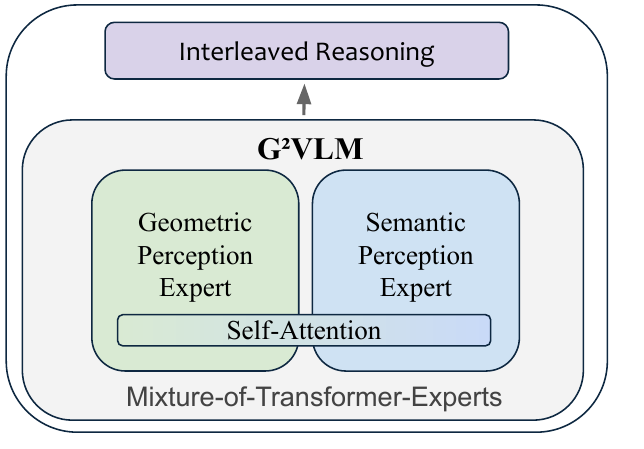}
\caption{Our model, \model, employs an architecture inspired by the two-streams hypothesis. It features two experts: a \textit{geometric perception} expert (our ``where pathway'') for visual geometry learning and a \textit{semantic perception} expert (our ``what pathway'') for multimodal understanding.}
 \label{fig:hypothesis}
 \vspace{-4mm}
\end{figure}

As illustrated in Figure~\ref{fig:hypothesis}, \model adopts a Mixture-of-Transformer-Experts (MoT)~\citep{liang2025mixtureoftransformerssparsescalablearchitecture,deng2025emerging} architecture that comprises two experts: one \textit{geometric perception} expert (our ``where pathway'') for visual geometry learning and one \textit{semantic perception} expert (our ``what pathway'') for multimodal understanding. These experts interact via shared self-attention, enabling the interplay between these two fundamental aspects to mutually improve one another. Our unified design also offers advantages in scaling. It learns to reason about 3D geometry from pure 2D image inputs, eliminating the reliance on difficult-to-collect 3D data (e.g., depth maps and camera poses) and allowing our framework to scale by leveraging abundant, in-the-wild multi-view images and videos for spatial understanding.

To train both experts effectively, \model employs a two-stage training strategy. In the first stage, the semantic perception expert is initialized from a frozen pretrained VLM (e.g., Qwen2-VL), while the geometric perception expert is trained from scratch on a large-scale 3D-annotated dataset to learn a geometry-rich representation. In the second stage, we unfreeze the semantic expert and jointly train it with the geometric expert on spatial understanding data. This allows it to effectively integrate the learned geometric features, further enhancing its spatial reasoning performance.



Through extensive experiments, \model demonstrates significant performance across a wide range of visual geometry and spatial reasoning tasks. On visual geometry tasks, \model achieves competitive results against state-of-the-art (SOTA) feed-forward 3D reconstruction models, such as VGGT~\citep{wang2025vggt}, across depth estimation, point estimation, and camera pose estimation tasks. Notably, it reduces the monocular depth estimation Absolute Relative Error from VGGT's 0.335 to 0.297 on the Sintel benchmark.
On spatial reasoning tasks, our model achieves the best results on SPAR-Bench among all existing works, surpassing GPT-4o by 18.5 points. Across all four spatial reasoning benchmarks, \model achieves results that are better than or comparable to those of much larger models, despite its relatively small 2B size. Furthermore, we conducts studies that confirms a positive interplay between geometric and semantic representations: as the geometric expert’s performance improves, it also leads to greater improvements in spatial reasoning.

\begin{figure*}[t]
  \centering
   \includegraphics[trim=0cm 24.5cm 4.5cm 0cm, clip, width=\linewidth]{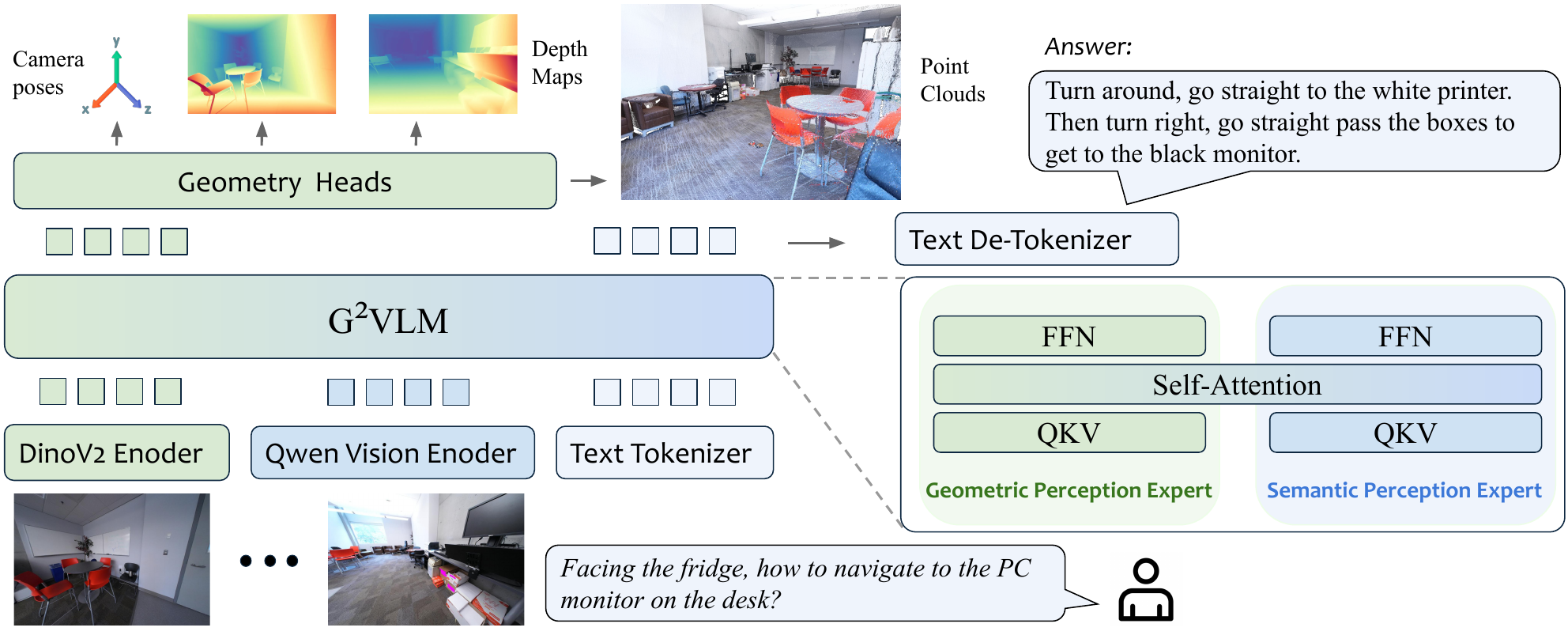}
  \caption{We present \model, a unified model that integrates both a geometric perception expert for 3D reconstruction and a semantic perception expert for multimodal understanding and spatial reasoning tasks. All tokens can do shared multi-modal self attention in each transformer block.}
 \label{fig:arch}
 \vspace{-4mm}
\end{figure*}

Our contributions can be summarized as follows: 
\begin{itemize}
[align=right,itemindent=0em,labelsep=2pt,labelwidth=1em,leftmargin=*,itemsep=0em]

\item We introduce \model, the first unified model that bridges spatial 3D  reconstruction and high-level spatial understanding in a single vision-language model.

\item \model employ a novel architecture with dedicated geometric and semantic perception experts, enabling the model to learn 3D geometry from 2D data and enhance spatial reasoning via shared self attention.

\item Experimental results demonstrate \model is proficient in both tasks, achieving competitive results against SOTA 3D reconstruction models and the best performance on most spatial reasoning benchmarks. 

\item By unifying a semantically strong VLM with low-level 3D vision tasks, we hope \model can serve as a strong baseline for the community and unlock more future applications, such as 3D scene editing.
\end{itemize}

\section{Related Works}
\noindent \textbf{VLMs as Unified Foundation Models} Recent advancements in Vision-Language Models (VLMs) have transformed multimodal research into a foundation-model paradigm. 
By combining visual understanding and language generation, VLMs demonstrates zero-shot generalizability across a wide range of multimodal tasks~\citep{video-chat, 3dllm,hu2023bliva, zhang2023llavagrounding}.
More recently, VLMs are being further unified to handle diverse tasks in an Any-to-Any input and output paradigm spanning audio, image, and video~\citep{li2025omniflowanytoanygenerationmultimodal, xie2025showo2improvednativeunified,ye2024xvilacrossmodalityalignmentlarge, lu2023unifiedio2scalingautoregressive}. One of the most successful unifications has been between multimodal understanding and generation, which has produced highly capable models~\citep{wang2024emu3nexttokenpredictionneed, xie2025showo2improvednativeunified, chen2025blip3ofamilyfullyopen, wang2025enhancedimagegenerationmultimodal}. Among these, Bagel~\citep{deng2025emerging} also adopts a Mixture-of-Transformer-Experts (MoT) architecture for multimodal understanding and image generation. While we adopt a similar design for our geometric and semantic perception experts, our work differs significantly. Our experts are trained for the vastly different tasks of visual geometry learning and spatial reasoning, which in turn necessitates distinct detailed model architecture design, pretraining objectives and joint training strategies from the one used by Bagel.



\noindent \textbf{Spatial Reasoning VLMs.}
Spatial understanding is a crucial step toward spatial intelligence, which is important for world understanding, robotics, and embodied AI. Recent progress in this area has led to the emergence of a wide range of spatial reasoning benchmarks~\citep{yang2025thinkingspacemultimodallarge, yang2025mmsibenchbenchmarkmultiimagespatial,lin2025ostbenchevaluatingcapabilitiesmllms, jia2025omnispatialcomprehensivespatialreasoning} and a corresponding rise in spatial VLMs~\cite{cheng2024spatialrgpt, ma2024spatialpin, chen2024spatialvlm, spaceqwen2025, ji2025robobrain}. Many of these works, including SpatialVLM and \citep{chen2024spatialvlm}, SpaceQwen \citep{spaceqwen2025}, adhere to the standard VLM design: they treat images as ``flat'' 2D data and rely on curated spatial reasoning datasets to achieve good performance.
Recognizing the limitations of 2D data priors, other recent works have noted the lack of geometric features in VLMs. For example, VLM-3R \citep{fan2025vlm} and Spatial MLLM \citep{wu2025spatialmllmboostingmllmcapabilities} both integrate a frozen geometry encoder (e.g., VGGT) as an extra representation and train the VLM to learn from these features. In contrast, our work directly integrates a geometric expert within our VLM, which offers a more natural alignment and unifies visual geometry prediction tasks with high-level spatial reasoning. Our model natively supports both geometric and semantic perception and can serve as a strong baseline to unlock future capabilities.

\noindent \textbf{Feedforward Visual Geometry.}
Recent works like DUSt3R \citep{wang2024dust3r} and MASt3R \citep{leroy2024grounding} alleviate the need for known camera parameters by utilizing Transformers~\citep{attention_is_all_need} to directly predict pixel-aligned 3D point maps. A subsequent wave of feed-forward visual geometry works, such as MV-DUSt3R+~\cite{Tang_2025_CVPR}, Cut3R~\cite{cut3r}, Fast3R~\citep{yang2025fast3r}, VGGT \citep{wang2025vggt}, and $\pi^3$ \citep{wang2025pi}, have further extended this approach to handle multi-view scenarios. These methods adopt a simple and efficient end-to-end inference to predict accurate 3D geometries, surpassing optimization-based pipelines. However, these methods remain focused on geometric reconstruction, often neglecting higher-level scene understanding and limiting their use in spatial reasoning tasks. 
Our work unifies visual geometry and multimodal reasoning within one framework by natively learning and predicting geometric features. This design inherits the geometric accuracy and efficiency of feed-forward architectures while enabling spatial comprehension for semantic downstream tasks.

\section{Unified Spatial Vision-Language Model }
\vspace{-1mm}
\label{sec: method}

We introduce \model, a unified geometry-grounded VLM that integrates spatial 3D reconstruction and spatial understanding. We present the model's architecture in $\S$~\ref{sec: model_arch} and detail the learning process for the geometric perception expert in $\S$~\ref{sec: vg_learning} and the semantic perception expert in $\S$~\ref{sec: spatial_learning}.


\subsection{Model Architecture}
\label{sec: model_arch}


As illustrated in Figure~\ref{fig:arch}, \model adopts a Mixture-of-Transformer-Experts (MoT) architecture~\cite{deng2025emerging} that consists of two transformer experts-one \textit{geometry perception expert} dedicated to visual geometry learning and one \textit{semantic perception expert} to multimodal understanding. Our model's input is a sequence $(I_i)_{i=1}^N$ of $N$ RGB images $I_i \in \mathbb{R}^{3\times H\times W}$, we present the detailed design for each expert as follows.

\noindent \textbf{Geometric Perception Expert.}
For geometric perception expert, we incorporate a DINOV2 vision encoder to inject low-level visual information to LLM which further reasons the 3D-aware feature through global attention. This process maps each image $I_i$ to LLM hidden states $h_i \in \mathbb{R}^{C \times d}$.
    These hidden states are decoded with 3D geometry heads for task-specific predictions, where the heads composed of a local point head, camera head and a global point head for stabilizing the training. All heads are designed as light-weight transformer decoders. The geometry heads is a function that maps the visual geometry hidden states to a corresponding set of 3D annotations:
    \begin{equation} 
f\left((h_i)_{i=1}^N\right)
=
\left(
T_i,
X_i,
\right)_{i=1}^N.
\end{equation}
where  $T_i \in SE(3) \subset \mathbb{R}^{4 \times 4}$ is the camera pose, $X_i \in \mathbb{R}^{H \times W \times 3}$ is the associated pixel-aligned 3D point map represented in its own camera coordinate system, each corresponding to the input image $I_i$.

\noindent \textbf{Semantic Perception Expert.} For semantic perception expert, we builds upon a pretrained VLM (Qwen2-VL-2B)~\cite{wang2024qwen2} as our foundation. We adopt the Qwen2 vision encoder which supports native dynamic resolution, along with the design of Multimodal Rotary Position Embedding (M-RoPE). It worth noting that our design can directly leverage off-the-shelf pretrained VLM which can be easily integrated with stronger VLMs. 


To better support visual geometry learning inside a VLM and narrow the representation gap to foster the mutual growth, we simplify some designs such as not incorporating register tokens~\cite{darcet2024visiontransformersneedregisters}, and use only global attention layers. To make our model further scalable, we also remove the camera token designs as in VGGT, and use permutation equivariant design as in $\pi^3$~\cite{wang2025pi}. These choices together makes the LLM directly utilize encoded visual features from both DINOV2 and semantic Qwen2 vision encoder in the same way without special priors. 



\subsection{Visual Geometry Learning}
\label{sec: vg_learning}

We train our model in two stages. First, we freeze the semantic perception expert which is loaded with Qwen2-VL weights, then we initialize the geometric perception expert from random parameters and train it from scratch using large-scale 3D annotated datasets. Following MoGe~\citep{wang2025moge} and $\pi^3$~\citep{wang2025pi}, our visual geometry (VG) loss function  $\mathcal L_{VG}$ is formulated as a weighted sum of the point reconstruction loss, the camera pose loss and the normal loss: 
\begin{equation}
\label{eq:vg_loss}
    \mathcal L_{VG} = \mathcal L_{\text{points}} + \lambda_{\text{cam}}\mathcal L_{\text{cam}} +
    \lambda_{\text{normal}}\mathcal L_{\text{normal}} 
\end{equation} 
where $\lambda$ is a pre-defined hyperparameter. Specifically, we describe each loss below.

The point cloud reconstruction loss, $\mathcal{L}_{\text{points}}$, is defined using the optimal scale factor $s^*$:
\vspace{-2mm}
\begin{equation}
\label{eq:points_loss}
\mathcal{L}_{\text{points}} = \frac{1}{3NHW}\sum_{i=1}^N \sum_{j=1}^{H \times W} \frac{1}{z_{i,j}} \| s^* \hat{\mathbf{x}}_{i,j} - \mathbf{x}_{i,j} \|_1 \end{equation}
where
$
s^* = \underset{s}{\arg\min} \sum_{i=1}^N \sum_{j=1}^{H \times W} \frac{1}{z_{i,j}} \| s \hat{\mathbf{x}}_{i,j} - \mathbf{x}_{i,j} \|_1
$.
Here, $\hat{\mathbf{x}}_{i,j} \in \mathbb{R}^3$ denotes the predicted 3D point at index $j$ of the point map $\hat{X}_i$. Similarly, $\mathbf{x}_{i,j}$ is its ground-truth counterpart in $X_i$. The term $z_{i,j}$ is the ground-truth depth, which is the z-component of $\mathbf{x}_{i,j}$ and is solved using the ROE solver proposed by~\cite{wang2025moge}.

The camera loss $\mathcal{L}_{\text{cam}}$ is a weighted sum of a rotation loss term and a translation loss term, averaged over all ordered view pairs where $i \neq j$:
\vspace{-2mm}
\begin{equation}
\label{eq:cam_loss}
    \mathcal{L}_{\text{cam}} = \frac{1}{N(N-1)} \sum_{i \neq j} \left( \mathcal{L}_{\text{rot}}(i, j) + \lambda_{trans} \mathcal{L}_{\text{trans}}(i, j) \right)\end{equation}
where $\lambda$ is a hyperparameter to balance the two terms. Specifically, the rotation loss minimizes the geodesic distance (angle) between the predicted relative rotation $\hat{R}_{i \leftarrow j}$ and its ground-truth target $R_{i \leftarrow j}$:
\vspace{-2mm}
\begin{equation}
\label{eq:rot_loss}
    \mathcal{L}_{\text{rot}}(i, j) = \arccos\left(\frac{\text{Tr}\left((R_{i \leftarrow j})^\top \hat{R}_{i \leftarrow j}\right) - 1}{2}\right)
\end{equation}
The translation loss is calculated using the Huber loss, $\mathcal{H}_\delta$, by comparing scaled prediction against the ground-truth relative translation, $t_{i \leftarrow j}$:
\vspace{-2mm}
\begin{equation}
\label{eq:trans_loss}
    \mathcal{L}_{\text{trans}}(i, j) = \mathcal{H}_\delta(s^*\hat{t}_{i \leftarrow j} - t_{i \leftarrow j})
\end{equation}

Finally, the normal loss $\mathcal{L}_{\text{normal}}$ encourages the reconstruction of locally smooth surfaces by minimizing the angle between the predicated normal vector $\hat{n}_{i,j}$ and their ground-truth counterparts $n_{i,j}$:
\begin{equation}
\label{eq:normal_loss}
\mathcal{L}_{\text{normal}} = \sum_{i=1}^N \sum_{j=1}^{H \times W} \arccos\left(\hat{n}_{i,j} \cdot n_{i,j}\right)
\end{equation}


\subsection{Spatial Reasoning Learning}
\label{sec: spatial_learning}

\begin{figure}[t]
  \centering
\includegraphics[width=0.9\linewidth]{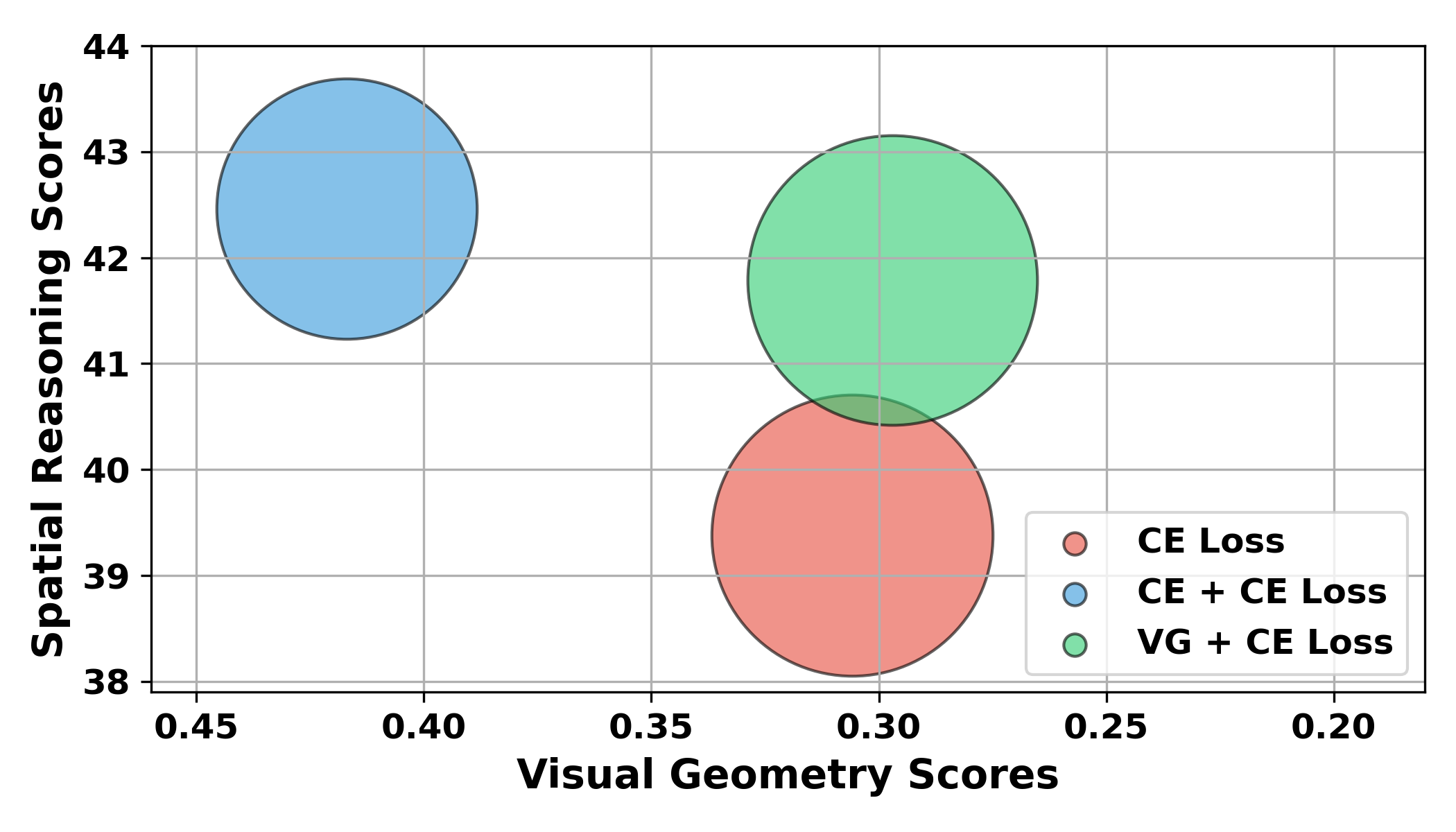}
\caption{Comparison of three different loss supervision mechanisms for the joint-training stage. Note that for visual geometry scores, lower is better. The \textit{VG + CE Loss} approach yields the best performance, demonstrating that combining visual geometry and spatial understanding supervision mutually benefits spatial reasoning tasks.}
 \label{fig:loss_curve}
 \vspace{-4mm}
\end{figure}

After pretraining the geometric perception expert, we perform a joint-training stage to leverage its learned geometric representations for spatial understanding tasks. The primary goal is to optimize the semantic perception expert to effectively utilize these geometric features via in-context learning and interleaved reasoning. This is achieved by optimizing the standard language modeling loss (cross-entropy). This process naturally raises several design choices. We explore three distinct joint-training strategies, where the semantic perception expert is, by default, optimized using a cross-entropy (CE) loss:
\begin{itemize}
[align=right,itemindent=0em,labelsep=2pt,labelwidth=1em,leftmargin=*,itemsep=0em]
    \item \textit{CE Loss Only}: Freeze the geometric perception expert, only the semantic perception expert is updated. This forces the model to learn to use the visual geometry features via in-context learning. This design also maintains the visual geometry performance unaltered.

    \item \textit{CE + CE Loss}: Optimize the geometric perception expert with CE loss. This fine-tunes the geometry features explicitly for spatial understanding.

    \item \textit{VG + CE Loss}: Optimize the geometric perception expert with both the CE loss and its visual geometry (VG) loss. This aims to adapt the expert for spatial understanding abilities while simultaneously maintaining its core pretrained geometry capabilities.
\end{itemize}
We conduct this study on the Scannet~\cite{dai2017scannet} dataset with semantic annotations.
As demonstrated in Figure~\ref{fig:loss_curve}, the \textit{VG + CE Loss} approach yields the strongest results, improving performance on both tasks. However, this method requires a large-scale 3D annotated dataset for the joint training stage, which limits its scalability.
Given this scalability constraint, we select the \textit{CE Loss Only} approach for our main \model. This method represents the best trade-off: it preserves the geometric expert's strong, pretrained geometry performance by freezing it, while also achieving good spatial reasoning capabilities by scaling with abundant video data.
Furthermore, the \textit{CE + CE Loss} variant was shown to be the most effective for optimizing spatial reasoning performance specifically. We therefore designate this specialized model as \textbf{\model-SR} (Spatial Reasoning) in this paper.


\noindent \textbf{Implementation Details.}
For the pre-training stage of geometric perception expert, we further divide it into two stage training. We first fix image resolution to 224x224 and use AdamW optimizer for $100$K iterations with a learning rate (lr) of 2e-4 using cosine scheduler. Then we further train at lr of 5e-4 for another $20$K steps with a higher resolution of 518x518, and apply randomized aspect ratio between 0.5 and 1.0.
Similar to VGGT, for every batch, we randomly sample $2$--$24$ frames from a random training scene.
The low-resolution pretraining runs on 32 A800 GPUs over 7 days and high-resolution runs on 64 A800 GPUs over 3 days.
For joint-training, we use AdamW optimizer for $16$K iterations with a lr of 2e-5 on 64 A800 GPUs over 3 days.
Throughout all training, we employ gradient norm clipping with a threshold of $1.0$ to ensure training stability and leverage bfloat16 precision and gradient checkpointing to improve GPU memory and computational efficiency.
We provide more training details in our supplementary materials. 

\noindent \textbf{Training Data.}
The model was trained using a large and diverse collection of datasets for the geometric perception expert, including:
ScanNet~\cite{dai2017scannet},
Co3Dv2~\cite{reizenstein2021common},
BlendMVS~\cite{yao2020blendedmvs},
DL3DV~\cite{ling2024dl3dv},
MegaDepth~\cite{li2018megadepth},
WildRGBD~\cite{xia2024rgbd},
TarTanAir~\cite{wang2020tartanair},
Taskonomy~\cite{zamir2018taskonomy},
ArkitScenes~\cite{baruch1arkitscenes},
HyperSim~\cite{hypersim},
Habitat~\cite{szot2021habitat},
ScanNetPP~\cite{yeshwanth2023scannet++},
GTA-SFM~\cite{wang2020flow},
Matrixcity~\cite{li2023matrixcity},
Aria Synthetic Environments~\cite{Pan_2023_ICCV},
Mapfree~\cite{arnold2022map},
and an internal synthetic indoor datasets. 
For joint training stage, we employ spatial understanding datasets such as SPAR-7M~\citep{zhang2025flatlandspaceteachingvisionlanguage}, the training sets of Omnispatial~\citep{jia2025omnispatialcomprehensivespatialreasoning}, Mindcube~\citep{yin2025spatialmentalmodelinglimited} and OST-Bench~\citep{lin2025ostbenchevaluatingcapabilitiesmllms} and general VQA dateset such as LLaVA-One-Vision~\cite{lillava}.


\begin{table*}[t]
\begin{subtable}{1\textwidth}
    \centering
 \small
 \renewcommand\tabcolsep{2.5pt} 
 \renewcommand\arraystretch{0.95} 
 \resizebox{0.94\linewidth}{!}{
  \begin{tabular}{c|cccc|cccc|ccc}
    \toprule
    \multicolumn{1}{c|}{\multirow{4}{*}{\textbf{Model}}} & \multicolumn{4}{c|}{\textbf{Depth Estimation}} & \multicolumn{4}{c|}{\textbf{Point Estimation}} & \multicolumn{3}{c}{\textbf{Camera Pose Estimation}}  \\
    \cmidrule(lr){2-5} \cmidrule(lr){6-9} \cmidrule(lr){10-12} 
    
    & \multicolumn{2}{c}{\textbf{Sintel}} & \multicolumn{2}{c}{\textbf{NYU-v2}} & \multicolumn{2}{c}{\textbf{ETH3D}} & \multicolumn{2}{c}{\textbf{7Scenes}} &  \multicolumn{3}{c}{\textbf{Co3Dv2}}\\
     \cmidrule(lr){2-3}\cmidrule(lr){4-5}\cmidrule(lr){6-7}\cmidrule(lr){8-9}\cmidrule(lr){10-12}
    
    & \header{Abs Rel$\downarrow$} & \header{$\delta<1.25\uparrow$} & \header{Abs Rel$\downarrow$} & \header{$\delta<1.25\uparrow$} & \header{Acc. $\downarrow$} & \header{Comp. $\downarrow$}& \header{Acc. $\downarrow$} & \header{Comp. $\downarrow$} & \header{RRA@30$\uparrow$} &\header{RTA@30$\uparrow$} & \header{AUC@30$\uparrow$} \\
    \midrule
    Fast3R~\cite{yang2025fast3r}       & 0.544 & 0.509 & 0.093 & 0.898  & 0.832 & 0.978 & 0.040 & 0.056 & 97.49 & 91.11& 73.43 \\
CUT3R~\cite{cut3r}        & 0.418 & 0.52  & 0.081 & 0.914    & 0.617 & 0.747 & 0.023 & 0.027 & 96.19 &92.69 & 75.82 \\
FLARE~\cite{zhang2025flare}        & 0.606 & 0.402 & 0.089 & 0.898 & 0.464 & 0.664 & 0.019 & 0.026 & 96.38& 93.76& 73.99 \\
VGGT~\cite{wang2025vggt}        & 0.335 & 0.599 & 0.056 & 0.951 & 0.28  & 0.305 & 0.022 & 0.026 & 98.96 & 97.13 & 88.59 \\
$\pi^3$~\cite{wang2025pi}         & 0.277 & 0.614 & 0.054 & 0.956  & 0.194 & 0.210  & 0.016 & 0.022 & 99.05 & 97.33 & 88.41 \\

    \midrule
     \rowcolor{black!10} 
    \model (Ours) & 0.297 & 0.589 & 0.062  & 0.954  & 0.414 & 0.309 & 0.046 & 0.029 & 97.91 & 95.20   & 74.81\\
     
    \bottomrule
    \end{tabular}
    }
\caption{\textbf{Visual Geometry tasks spanning Depth Estimation, Point Estimation, and Camera Pose Estimation}. Our model, \model, demonstrate comparable performance against SOTA feed-forward 3D recontruction methods. 
}
\label{tab: vg}
\end{subtable}

\begin{subtable}{1\textwidth}
\centering
 \renewcommand\tabcolsep{2.5pt} 
 \renewcommand\arraystretch{0.95} 
 \resizebox{0.98\linewidth}{!}{
    \begin{tabular}{c|cccc|cccc|cccc|ccc}

    \toprule
    \multicolumn{1}{c|}{\multirow{4}{*}{\textbf{Model}}}   &\multicolumn{4}{c|}{\textbf{SPAR-Bench}} & \multicolumn{4}{c|}{\textbf{MindCube}} & \multicolumn{4}{c|}{{\textbf{OST-Bench$^*$}}}  & \multicolumn{3}{c}{{\textbf{OmniSpatial$^*$}}}  \\
      \cmidrule(lr){2-5}  \cmidrule(lr){6-9} \cmidrule(lr){10-13} \cmidrule(lr){14-16}
      & \header{Avg.} & \header{Low}  & \header{Medium} & \header{High}  & \header{Avg.} & \header{Rotation}  & \header{Among} & \header{Around}  & \header{Avg.} & \header{A. State}  & \header{A. Info} & \header{AO.}  & \header{Avg.} & \header{SI}  & \header{PT}  \\
    \midrule 
    \multicolumn{16}{l}{\hfill \textit{Proprietary Models} } \\
    \midrule 
    GPT-4o~\cite{openai2024gpt4o} &36.39 & 29.25& 24.93& 45.11& 38.81& 32.65 &40.17 &29.16 & 53.05 &37.58 &77.66&41.93 &50.74 &59.31 &46.16\\
    Claude-3.7-Sonnet~\cite{anthropic2025claude37sonnet} & 21.77 & 25.43 & 7.33 & 23.33 &40.50 &41.33 &40.00&39.50 &-&-&-&- &48.76 &60.07 &42.71 \\
    Claude-4-Sonnet-20250514~\cite{anthropic2025claude4sonnet} & - & - & - & - & 44.75 & 48.42 & 44.21 & 47.62 & - & - & - & - & - & -  & -\\
    \midrule
    \multicolumn{16}{l}{\hfill \textit{Open-source Models} } \\
    \midrule 
    LLaVA-v1.5-7B~\cite{liu2024improved} & 23.65 & 10.85 & \underline{26.50} & 34.09 &-&-&-&-&-&-&-&-&37.44 &35.11 &38.69 \\
    LLaVA-Video-7B ~\cite{zhang2024videoinstructiontuningsynthetic} & 32.33 & 23.55 & 24.83 & 42.62 &\underline{41.96} &35.71 & \underline{43.55} &30.12&44.11 &36.80 &64.49&32.51&-&-&- \\
    LLaVA-OneVision-7B~\cite{lillava} &31.20 & 21.79 & 26.12 & 40.14 & \textbf{47.43} & 36.45 & \textbf{48.42} & \textbf{44.09} &43.93 &28.81 &61.49 &\underline{37.14} &37.17 &35.00 &38.32\\
    Qwen2-VL-7B~\cite{wang2024qwen2} & 30.74 & 27.52 & 20.44 & 37.03 &31.50&35.50&31.00&25.75&36.56&32.49&47.60&32.49&-&-&-\\
    InternVL2.5-8B~\cite{chen2024expanding} & \underline{36.28} & \underline{29.46} & \textbf{31.88} & \underline{43.80} & 18.68 &36.45 &18.20 &13.11 &\underline{49.56} &\underline{45.38} &\underline{65.12} &\textbf{39.51} &-&-&-\\
    Qwen2.5-VL-3B~\cite{bai2025qwen2} & 29.39 & 26.69 & 24.87 & 33.29 & 33.21 &37.37 & 33.26& 30.34 &31.17&26.94&49.35&18.87&39.98 &44.27 &37.68\\
    Qwen2.5-VL-7B~\cite{bai2025qwen2} & 33.07 & 28.75 & 22.97 & 40.27 & 29.26 & \underline{38.76} &29.50 &21.35 &44.22 &42.19 &56.12 &34.21&\underline{41.83} &\underline{46.20} &\underline{39.50} \\
    Qwen2.5-VL-72B~\cite{bai2025qwen2} & \textbf{39.40} & \textbf{35.35} & 23.05 & \textbf{48.44} &37.25 &\textbf{41.17} &32.50 &\underline{30.50} & \textbf{50.54} &\textbf{48.69} &\textbf{71.59} &31.09& \textbf{47.13} &\textbf{54.80} &\textbf{43.03} \\
    \midrule
     \multicolumn{16}{l}{\hfill \textit{Spatial Expert Models} } \\
    \midrule 
    SpaceMantis-13B~\citep{remyxai_spacemantis13b_2025} & 28.93 & 23.56 &
    23.27 &35.60&22.81 &37.65 &21.26 &29.32 & -&-&-& - &38.84 &35.93 &40.40 \\
    Spatial-MLLM-7B~\cite{wu2025spatialmllmboostingmllmcapabilities}  &32.15 & 29.88&20.30 &38.13 &  32.06& 38.39& 20.92& \underline{32.82} &\underline{26.72} &22.95 &42.19 &\underline{20.97} &44.94&47.33&43.67\\
    SpaceQwen2.5-VL-3B ~\cite{spaceqwen2025} &24.24 &14.46 & 24.00& 33.01& 33.28 & 38.02 & 33.71 & 26.32 &26.61&{26.69} &\underline{43.05}&13.82 &42.63 &41.00 &43.49\\
    VLM3R-7B~\citep{fan2025vlm} & \underline{43.21} & 39.78 & 28.43 &\underline{51.18} &\underline{42.09}& 	36.73 	&\textbf{44.22} 	&24.45& -& - & -& - &\underline{48.08}&\textbf{55.33}&44.21\\
    \midrule
    Qwen2-VL-2B (base model) & 24.60 & 19.43 & 27.55 & 28.22 &  37.83 & 34.50 & \underline{41.16} &  32.25 & 22.59 & \underline{30.43} & 26.85 & 17.94 &43.90 & {49.00} & 41.18
\\
\rowcolor{black!10} 
    \model-2B (Ours)  &41.66 & \underline{56.49} & \underline{32.60} &31.49 &  38.83 & \underline{47.00}&  26.50 & 32.75 & 24.60 &25.32 & 36.23 &17.18 & {44.13} & 42.67 &\underline{44.92}\\
   \rowcolor{black!10} 
    \model-SR-2B (Ours)  &\textbf{54.87} & \textbf{59.99}  & \textbf{36.27} & \textbf{56.51} & \textbf{48.33} & \textbf{53.17} & 31.50 &\textbf{49.50} & \textbf{37.02} &\textbf{30.73} & \textbf{45.54} &\textbf{31.42} & \textbf{50.41} & \underline{52.67} &\textbf{49.20}\\
    
    \bottomrule
    \end{tabular}
    }
\caption{\textbf{Spatial Understanding and Reasoning Results.}
Higher scores are better. We \textbf{bold} the best and \underline{underline} the second-best results for open-source and spatial expert models, separately.
OST-Bench$^*$ denotes a subset with $\le$15 input frames.
OmniSpatial$^*$ denotes evaluation on its two main categories: Spatial Interaction (SI) and Perspective Taking (PT). \model-SR achieves the best results on most tasks, while \model also demonstrates significant improvements over the base model on all tasks.
}
\label{tab: und}
\end{subtable}

\caption{Comparison with mainstream feed-forward 3D reconstruction methods on visual geometry tasks and with representative VLMs on spatial understanding and reasoning tasks. Our model demonstrates proficient performance in both aspects of spatial tasks, demonstrating its universality and effectiveness.}
\vspace{-4mm}
\label{tab:main table}
\end{table*}

\vspace{-2mm}

\section{Experiments}
\label{sec: results}

We evaluate our method across a wide range of spatial tasks. For visual geometry, we evaluate on monocular depth estimation, point map estimation, and camera pose estimation tasks (§~\ref{sec: results_vg}). For spatial understanding and reasoning, we evaluate our model on comprehensive benchmarks, including SPAR-Bench~\citep{zhang2025flatlandspaceteachingvisionlanguage}, OmniSpatial~\cite{jia2025omnispatialcomprehensivespatialreasoning}, MindCube~\cite{yin2025spatialmentalmodelinglimited} (spatial mental modeling), and OST-Bench~\cite{lin2025ostbenchevaluatingcapabilitiesmllms} (online spatio-temporal scene understanding) (§~\ref{sec: results_und}).
Across the visual geometry tasks, our model achieves performance comparable to existing feed-forward 3D reconstruction methods. Furthermore, it achieves the best or comparable performance results on most spatial understanding and reasoning tasks. We also discuss our design choices and validate them through a detailed ablation study in $\S$~\ref{sec: discussion}.

\begin{figure*}[t]
\centering

\includegraphics[trim=0cm 24cm 2.3cm 0cm, clip, width=\linewidth]{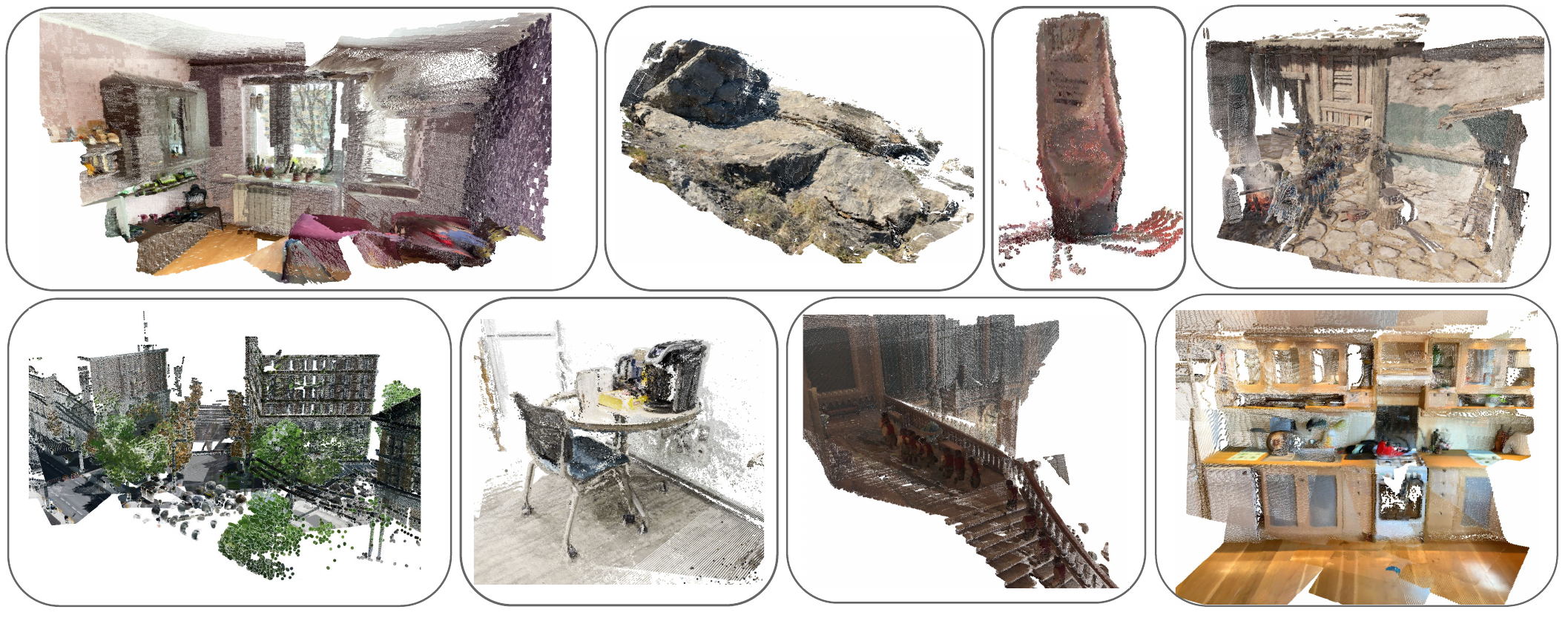}

\caption{Qualitative results of our model. \model effectively reconstructs a diverse set of open-domain images, spanning object-level, structure-level, indoor, and outdoor scenes, including both dynamic and static content.}
\label{fig:qualitative}
\vspace{-5mm}
\end{figure*}

\subsection{Visual Geometry Results}
\label{sec: results_vg}

\noindent \textbf{Monocular Depth Estimation.}
Following the methodology of~\cite{wang2024dust3r,cut3r,wang2025pi}, we evaluate our method on monocular depth estimation task using the Sintel~\cite{bozic2021transformerfusion} and NYU-V2~\cite{silberman2012indoor} datasets. We report the Absolute Relative Error (Abs Rel) and the prediction accuracy at a threshold of $\delta<1.25$. 
As demonstrated in Table~\ref{tab: vg}, our method achieves on-par performance with SOTA multi-frame feed-forward reconstruction approaches, such as VGGT and $\pi^3$.

\noindent \textbf{Point Map Estimation.} Following the evaluation settings in~\cite{cut3r, wang2025pi}, we evaluate the quality of reconstructed multi-view point maps on the 7-Scenes~\cite{shotton2013scene} and ETH3D~\cite{schops2017multi} datasets. We sample keyframes using a stride of 40 for 7-Scenes and every 5 images for ETH3D. Predicted point maps are aligned to the ground truth using the Umeyama algorithm for a coarse Sim(3) alignment, followed by refinement with the Iterative Closest Point (ICP) algorithm.
Consistent with prior works~\cite{azinovic2022neural,wang2024dust3r,wang2024spann3r,cut3r}, we report Accuracy (Acc.) and Completion (Comp.) in Table~\ref{tab: vg}. These results demonstrate that our method achieves on-par performance with VGGT in completion and comparable results in accuracy. This confirms our model's effectiveness, despite that it uses a simpler attention mechanism. 


\noindent \textbf{Camera Pose Estimation.} Following prior work~\cite{wang2024dust3r,wang2025vggt}, we evaluate predicted camera poses on the Co3Dv2~\cite{reizenstein2021common} dataset, which features over 1000 test sequences, using angular accuracy. For each sequence, we randomly sample 10 images, form all possible pairs, and compute the angular errors of the relative rotation and translation vectors. This process yields the Relative Rotation Accuracy (RRA) and Relative Translation Accuracy (RTA) at a given threshold (e.g., RRA@30 for 30 degrees). The Area Under the Curve (AUC) of the min(RRA,RTA)-threshold curve serves as a unified metric.
As shown in Table~\ref{tab: vg}, our method achieves on-par performance on the RRA and RTA metrics and comparable results on the AUC metric when compared with SOTA models. These results underscore our model's strong capabilities, particularly since it does not use camera tokens (like VGGT) which provides a strong camera pose prior or require fine-tuning from pre-trained weights (like $\pi^3$).

We further present qualitative results of \model in Figure~\ref{fig:qualitative}. Our model generates high-quality predictions and generalizes well across a diverse set of open-domain images, spanning object-level, structure-level, indoor, and outdoor scenes, including both dynamic and static content.

\subsection{Spatial Understanding \& Reasoning Results}
\label{sec: results_und}

We evaluate our model on comprehensive spatial understanding and reasoning benchmarks, including SPAR-Bench~\citep{zhang2025flatlandspaceteachingvisionlanguage}, OmniSpatial~\citep{jia2025omnispatialcomprehensivespatialreasoning}, MindCube~\citep{yin2025spatialmentalmodelinglimited} (spatial mental modeling), and OST-Bench~\citep{lin2025ostbenchevaluatingcapabilitiesmllms} (online spatio-temporal scene understanding). As demonstrated in Table~\ref{tab: und}, we compare our model against a wide range of methods, including proprietary, open-source, and spatial-expert models.

Across all tasks, both \model and \model-SR demonstrate a significant improvement over our base model, Qwen2-VL-2B, illustrating the effectiveness of our approach. Notably, \model-SR achieves the best results among all existing works, surpassing the proprietary GPT-4o by 18.48 points on SPAR-Bench. When compared with spatial expert models, \model-SR also achieves the best results on all four spatial benchmarks, despite its relatively small 2B size.

Comparing with open-source models, \model-SR achieves the best results on SPAR-Bench, Mindcube, and OmniSpatial. On OST-Bench, the much larger Qwen2.5-VL-72B demonstrates the best results. This suggests a trend that online spatio-temporal scene understanding requires models to store significant knowledge, favoring larger architectures. It is worth noting, however, that \model-SR still outperforms spatial expert models of similar or larger sizes on OST-Bench, further demonstrating the effectiveness of our architecture. We leave the scaling of our model to future work, as this is a promising direction to unlock even stronger performance.

\begin{figure}[t]
  \centering
\includegraphics[width=\linewidth]{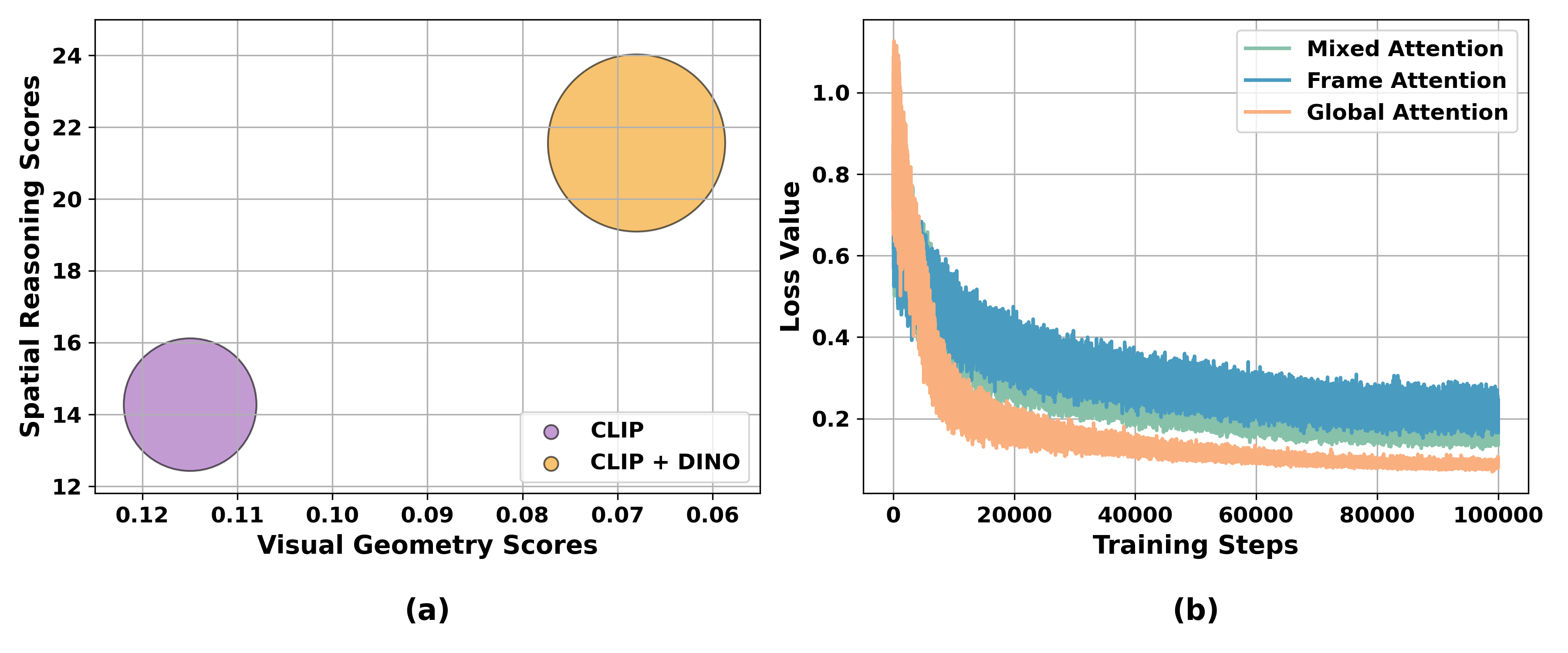}
  \caption{Experimental study results. (a) The dual encoder design, with both a semantic-rich CLIP encoder and a low-level vision DINO encoder, yields the best performance on both visual geometry and spatial understanding tasks. (b) Training loss curves for three different attention mechanisms during geometric perception expert training; global attention is consistently the best variant.}
 \label{fig:exp_study}
 \vspace{-4mm}
\end{figure}

\subsection{Discussions and Ablation Study} 
\label{sec: discussion}

In our work, we adopt a dual-expert design comprising two distinct experts: one for geometric perception and one for semantic perception. This naturally raises several research questions about the optimal design.

\noindent \textbf{Encoder Design: Single vs. Dual?} Our first question is: Do the two experts require separate vision encoders, or can they share one? Can benefits be gained from pure visual geometry learning? VLMs typically adopt vision encoders trained with language-contrastive loss (e.g., CLIP~\citep{clip}, SigLIP~\citep{zhai2023sigmoid}), as their strong semantic capabilities provide a natural basis for visual understanding. In contrast, visual geometry models often utilize self-supervised encoders like DINO~\citep{caron2021emerging}, which is well-suited for low-level vision tasks such as 3D reconstruction. To bridge these two domains, we compare a single-encoder design (using CLIP for both tasks) with a dual-encoder design (using DINO for visual geometry and CLIP for multimodal understanding). As demonstrated in Figure~\ref{fig:exp_study}(a), the dual-encoder design yields the best results for both tasks. While DINO's suitability for low-level tasks is intuitive, we found it also significantly benefits spatial understanding. This indicates that DINO incorporates visual information that complements the multimodal vision encoder, ultimately enhancing performance on spatial reasoning tasks.




\noindent \textbf{Attention Mechanism for Geometric Perception Expert.} Our second question is: What is the most effective attention mechanism for the geometry expert?
Existing state-of-the-art visual geometry models such as MapAnything~\citep{keetha2025mapanything}, VGGT~\citep{wang2025vggt}, and $\pi^3$~\citep{wang2025pi} adopt transformers with alternating-attention layers, which switch between frame-wise and global attention. 
However, this architecture is not directly compatible with modern LLM frameworks that apply consistent attention masks at every layer. Therefore, we explore three mask-based attention variants: 1) frame attention, where the mask is applied to each frame individually; 2) global attention, where the mask is applied to the entire sequence of all frames; and 3) mixed attention, which randomly alternates between masks to encourage the model to learn both local geometry and global correspondence. As illustrated in Figure~\ref{fig:exp_study}(b), the global attention variant consistently outperforms the other designs, demonstrating its effectiveness within an LLM-compatible framework.

\begin{table}
\centering
\small
\renewcommand\tabcolsep{1.pt} 
\renewcommand\arraystretch{0.95} 
 \resizebox{0.95\linewidth}{!}{
\begin{tabular}{l|cccc} 
\toprule
\multicolumn{1}{l|}{\multirow{3}{*}{\textbf{Model}}} & \multicolumn{4}{c}{\textbf{SPAR-Bench}} \\
\cmidrule(lr){2-5} 
& \header{Avg.} & \header{Low} & \header{Medium} & \header{High} \\
\midrule

Qwen2-VL-2B (base model) & 24.60 & 19.43 & 27.55 & 28.22 \\
Qwen2-VL-2B (finetuned) & 48.93 & 56.31 & 24.82 & 50.42 \\
\model-SR (Frame-Att. in GP) &52.34 & 58.23 & 34.47 & 53.81\\
\model-SR (Mixed-Att. in GP)&53.64 & 59.16 & 35.33 & 55.16\\

\rowcolor{black!10}
\model-SR (Ours) &\textbf{54.87} & \textbf{59.99}  & \textbf{36.27} & \textbf{56.51}  \\

\bottomrule
\end{tabular}
}

\caption{Ablation study on the design choices for \model. GP denotes the geometric perception expert. Our results validate the superiority of our approach over the baselines. Notably, it confirms a positive interplay between geometric and semantic representations: as the geometric expert’s performance improves, it also leads to greater improvements in spatial reasoning.}
\vspace{-4mm}
\label{tab:ablation}
\end{table}


\noindent \textbf{Interplay Between Geometry and Reasoning.} Following this, we investigated if better geometric features lead to greater performance improvements on spatial reasoning tasks. As shown in Table~\ref{tab:ablation}, the global attention design (which is the best for geometry) also yields the strongest spatial reasoning results. This confirms a positive interplay between the two representations: as the geometric expert's performance improves, it also leads to greater improvements in spatial reasoning.

\noindent \textbf{Impact of Geometry Pretraining.} Finally, we conduct an ablation to isolate the overall effect of our geometry expert. We compare our full model to a baseline variant that was finetuned only on the spatial understanding data. As shown in Table~\ref{tab:ablation}, our model significantly outperforms this finetuned-only variant, demonstrating that the learned visual geometry representations are essential to our model's effectiveness.

\section{Conclusion}
\vspace{-2mm}
\label{sec: conclusion}
In this work, we introduce \model, a unified vision-language model that bridges two fundamental aspects of spatial intelligence: 3D reconstruction and spatial understanding. \model natively leverages learned 3D visual geometry features to directly predict 3D attributes and enhance spatial reasoning tasks via in-context learning and interleaved reasoning. Experimental results demonstrate \model is proficient in both tasks, achieving competitive results against SOTA 3D reconstruction models and the best performance on most spatial reasoning benchmarks. 
While our model exhibits strong generalization abilities in both visual geometry and spatial reasoning, one potential limitation is training instability with large-scale models. This challenge requires advanced optimization techniques, careful data curation, and significant computational resources. However, the primary advantage of our approach is its universality and effectiveness. It can serve as a strong baseline for the community to unlock future opportunities in semantic spatial tasks, spanning both high-level and low-level applications.




{
    \small
    \bibliographystyle{ieeenat_fullname}
    \bibliography{main}
}


\clearpage
\setcounter{page}{1}
\maketitlesupplementary
\setcounter{page}{1}
\setcounter{section}{0} 
\renewcommand{\thesection}{\Alph{section}} %

\section{Architecture Details}

We employ a pretrained DINOv2 encoder. Unlike VGGT~\citep{wang2025vggt} or $\pi^3$~\citep{wang2025pi}, where DINO features directly match the geometry transformer's hidden dimension, our architecture adopt an additional feature alignment step. Because we initialize the geometric perception expert to match the dimensions of the underlying LLM (Qwen2-VL-2B), we use a linear projection layer to map the DINO features into the expert's input space. This approach is very commonly adopted in VLM literature. Both the geometric expert and sematic expert comprises 28 layers, which mirroring the Qwen2-VL-2B architecture, and utilizes global attention. 

In contrast to VGGT, which relies on a computationally intensive DPT head that aggregates multi-scale features, we adopt lightweight transformer-based geometry heads which is similar to $\pi^3$. The decoders for camera poses, local point maps, and global point maps, share the same trasformer architecture but do not share weights. This architecture is a lightweight, 5-layer transformer that applies self-attention exclusively to the features of each individual image.  Following the decoder, the output heads vary by task. The heads for local point maps consist of a simple MLP followed by a pixel shuffle operation. For camera poses, the head is adapted from Reloc3r \cite{dong2025reloc3r} and Pi3~\cite{wang2025pi} and uses an MLP, average pooling, and another MLP. The rotation is initially predicted in a 9D representation \cite{levinson2020analysis} and is then converted to a 3×3 rotation matrix via SVD orthogonalization. As we mentioned in Sec 3.1, the global point head serves solely to stabilize training and is excluded during inference.

\section{Training Details}
\label{appendix: train details}

For the pre-training stage of geometric perception expert, we further divide it into two stage training. We first fix image resolution to 224x224 and use AdamW optimizer for $100$K iterations with a learning rate (lr) of 2e-4 using cosine scheduler. Then we further train at lr of 5e-4 for another $20$K steps with a higher resolution of 518x518, and apply randomized aspect ratio between 0.5 and 1.0. 
Similar to VGGT, for every batch, we randomly sample $2$--$24$ frames from a random training scene.
For our visual geometry loss function, we set the weights for each component as follows: $\lambda_{\text{normal}} = 1.0$, $\lambda_{\text{cam}} = 0.2$, and $\lambda_{\text{trans}} = 200.0$. The implementation of our normal loss follows that of MoGe~\cite{wang2025moge} and $\pi^3$~\cite{wang2025pi}, and the resolution for aligning the local point map loss is set to 4096.
The low-resolution pretraining runs on 32 A800 GPUs over 7 days and high-resolution runs on 64 A800 GPUs over 3 days. For visual geometry learning, we clip the loss that is greater than 10 and smooth it to 0 to avoid training instabilities. The loss spikes are due to noisy large 3D annotation data and further data cleaning efforts can help minimize these phenomena.  

For joint-training, we use AdamW optimizer and cosine scheduler for $16$K iterations with a lr of 2e-5 on 64 A800 GPUs over 3 days. We do not apply loss clipping here.
Throughout all training, we employ gradient norm clipping with a threshold of $1.0$ to ensure training stability and leverage bfloat16 precision and gradient checkpointing to improve GPU memory and computational efficiency.

\section{More results}

We evaluate our model on the SPAR-Bench which is a comprehensive spatial reasoning benchmark. Here we present the detailed results of each sub-task in SPAR-Bench in Table~\ref{tab:sparbench_res_detail}. Our model, \model-SR, demonstrates the best performance consistently across all tasks. Notably, it surpasses human performance in the low category.

\renewcommand{\arraystretch}{1.25} 
\setlength{\tabcolsep}{3pt} 
\begin{table*}[ht]
\footnotesize
    \centering
    \small
    \resizebox{1.0\textwidth}{!}{
    \begin{tabular}{l| c c | c c c c c c c c c| c c c c| c c c c c c c c c c}
        \textbf{Methods} & \rotatebox{75}{\textbf{Rank}} & \normalsize\rotatebox{75}{\textbf{Avg.}} & \normalsize\rotatebox{75}{\textbf{Low}} & 
       \scriptsize \rotatebox{75}{Depth-OC} & \scriptsize \rotatebox{75}{Depth-OC-MV} & \scriptsize \rotatebox{75}{Depth-OO} & \scriptsize \rotatebox{75}{Depth-OO-MV} & \scriptsize \rotatebox{75}{Dist-OC} & \scriptsize \rotatebox{75}{Dist-OC-MV} & \scriptsize \rotatebox{75}{Dist-OO} & \scriptsize \rotatebox{75}{Dist-OO-MV} &
       \normalsize \rotatebox{75}{\textbf{Medium}} &
       \scriptsize \rotatebox{75}{PosMatch} & \scriptsize \rotatebox{75}{CamMotion} & \scriptsize \rotatebox{75}{ViewChgI} &
       \normalsize \rotatebox{75}{\textbf{High}} & \scriptsize \rotatebox{75}{DistI-OO} & \scriptsize \rotatebox{75}{DistI-OO-MV} & \scriptsize \rotatebox{75}{ObjRel-OC-MV} & \scriptsize \rotatebox{75}{ObjRel-OO} & \scriptsize \rotatebox{75}{ObjRel-OO-MV} & \scriptsize \rotatebox{75}{SpImag-OC} & \scriptsize \rotatebox{75}{SpImag-OC-MV} & \scriptsize \rotatebox{75}{SpImag-OO} & \scriptsize \rotatebox{75}{SpImag-OO-MV} \\
        \hline
        \textbf{Baseline}& & & & & & & & & & & & & & & & & & & & & & &&& \\
        Chance Level (Random) & -& -& -& -& -& -& -& -& -& -& -&- &22.65&24.50& -& 25.09 &23.82& 22.02 & 31.25 & 25.27 & 22.16 & 25.81 & 24.42& 24.17 & 26.89\\
        Chance Level (Frequency) & - & 32.74 & 31.19 & 43.09 & 43.51 & 17.38 & 13.05 & 41.90 & 30.99 & 27.40 & 32.17 & 38.25 &29.01 & 26.75 & 59.00 &32.29& 52.94 & 50.60 & 28.25 & 26.92 & 26.59 & 26.34 & 26.74 & 26.49 & 25.77 \\
        \hline
        \textbf{SPAR-Bench \textit{(tiny) API}}& & & & & & & & & & & & & & & & & & & & & & &&&\\
        Human Level & 1& \cellcolor{lightred}67.27&\cellcolor{lightred} 55.31& 72.75& 74.25& 28.75& 36.25&78.25& 52.25& 66.5& 33.50& \cellcolor{lightred}72.32 &92& 64& 60.97& \cellcolor{lightred}76.22&80& 94& 70& 92& 80& 78& 82& 50&60\\
        GPT-4o~\cite{achiam2023gpt} & 3 & \cellcolor{lightyellow}36.39& 29.25 & 53.80& 45.00& 15.00& 13.60& 37.40& 34.40& 23.40& 24.40& \cellcolor{lightorange}24.93&30& 16& 28.80&\cellcolor{lightyellow}45.11& 64& 64& 58& 46& 46& 32& 44& 30&22\\
        Claude-3.7-Sonnet~\cite{anthropic2025claude37sonnet} & 5 & 21.77 & 25.43& 41.00 & 45.40 & 11.20 & 12.20& 42.60& 19.60& 26.00 & 5.40 & 7.33&16 & 6 & 0.00 & 23.33&40& 48 & 22 & 36 & 14 & 12 & 20 & 6& 12 \\
        Qwen2-VL-72B~\cite{wang2024qwen2} & 4 & 35.62 & \cellcolor{lightyellow}35.28& 45.40 & 49.80 & 13.80 & 10.00&54.60 & 49.40& 36.80 & 22.40 & \cellcolor{lightyellow}23.39&42 & 18 & 10.16&40.00&60 & 68& 50 & 38 & 44 & 18 & 28 & 18& 36 \\
        Qwen2.5-VL-72B~\cite{bai2025qwen2} & 2 &  \cellcolor{lightorange}39.40 & \cellcolor{lightorange}35.35 & 53.20 & 46.80 & 17.80 & 29.00& 49.60& 57.40&  14.40 & 14.60 & 23.05 & 40& 16 & 13.16 & \cellcolor{lightorange}48.44 & 74& 74& 60&56 &50 & 20& 34 & 24& 44 \\
        \hline
        \textbf{SPAR-Bench \textit{(full)}}& & & & & & & & & & & & & & & & & & & & & & &&&\\

        LLaVA-Video-7B & 5 &32.33 & 23.55& 26 & 37.4& 26.9 & 12 & 16.4 & 16.3 & 26.9 & 26.5 & 24.83 & 35.4 & 30.8 & 8.3  & 42.62 & 58.2 & 54.2 & 53.8 & 43.1 & 38.8 & 34.9 & 36.6 & 26.5 & 37.5 \\
        
        InternVL2-2B~\cite{chen2024far} & 15& 28.06& 21.74& 18.06& 24.81& 23.20& 20.97& 19.47& 19.95& 26.83& 20.61& 22.83&39.69& 23.00& 5.81& 35.42&51.18& 55.95& 46.00& 31.59& 23.82& 36.02& 34.30& 17.55&22.41\\

        InternVL2-4B~\cite{chen2024far} & 7& 32.01 & 28.94& 23.94 & 27.22 & 20.00 & 18.12 & 42.57 & 40.16 & 31.29 & 28.18 & 29.16&49.87 & 21.00 & 16.62 & 35.70&56.76 & 55.36 & 40.25 & 36.81 & 25.21 & 28.76 & 32.27 & 21.19 & 24.65\\

        InternVL2.5-2B~\cite{chen2024expanding} & 11& 30.14 & 25.79& 39.67 & 39.72 & 12.12 & 15.03 & 30.94 & 29.59 & 20.22 & 19.02 & 22.93&37.91 & 24.25 & 6.64 & 36.41&51.47 & 56.85 & 50.25 & 33.79 & 24.10 & 27.15 & 35.17 & 26.49 & 22.41\\

        InternVL2.5-4B~\cite{chen2024expanding} & 10& 30.55& 25.66& 29.06& 32.97& 21.77& 16.83& 20.84& 26.85& 28.13& 28.79& \cellcolor{lightyellow}29.75&47.07& 33.25& 8.92& 35.16&54.12& 58.93& 35.50& 29.67& 34.63& 24.73& 31.39& 19.21&28.29\\

        InternVL2.5-8B~\cite{chen2024expanding} & 3& \cellcolor{lightyellow}36.28 & 29.46& 25.78 & 29.31 & 23.79 & 18.76 & 46.82 & 42.68 & 22.62 & 25.89 & \cellcolor{lightorange}31.88 &61.32 & 28.00 & 6.32 &\cellcolor{lightyellow}43.80 &59.71 & 56.85 & 51.75 & 44.23 & 41.55 & 36.56 & 41.57 & 22.52 & 39.50\\

        LLaVA-OV-0.5B~\cite{lillava} & 12& 29.48 & \cellcolor{lightyellow}30.14& 49.22 & 42.72 & 18.04 & 14.92 & 31.48 & 25.67 & 28.98 & 30.10 & 15.89&24.43 & 21.75 & 1.50 & 33.42&50.88 & 50.00 & 32.00 & 27.75 & 26.04 & 30.91 & 34.01 & 24.50 & 24.65\\

        LLaVA-OV-7B~\cite{lillava} & 8& 31.20 & 21.79& 30.33 & 26.94 & 18.58 & 13.87 & 10.43 & 13.64 & 31.24 & 29.29 &26.13 &38.68 & 30.25 & 9.47 &40.14 &56.47 & 55.06 & 37.25 & 48.63 & 38.23 & 30.38 & 33.72 & 26.49 & 35.01\\

        Qwen2-VL-2b~\cite{wang2024qwen2} & 16& 24.60 & 19.43& 38.03 & 40.63 & 18.84 & 14.09 & 7.81 & 7.07 & 17.82 & 11.14 & 27.55&26.21 & 25.25 & 31.20 &  28.22&54.12 & 49.11 & 21.75 & 25.27 & 12.47 & 23.92 & 27.62 & 24.83 & 14.85\\

        Qwen2-VL-7b~\cite{wang2024qwen2} & 9& 30.74 & 27.52& 35.97 & 35.22 & 20.83 & 12.88 & 28.68 & 29.95 & 28.21 & 28.45 & 20.44&35.37 & 20.25 & 5.69 & 37.03&59.71 & 52.38 & 30.25 & 38.46 & 41.00 & 22.04 & 28.49 & 22.52 & 38.38\\

        Qwen2.5-VL-3B~\cite{bai2025qwen2} & 13 & 29.39  & 26.69  & 31.7 &34.2 & 32.1 & 17.5 & 18.4 & 22.7 & 32.1 & 24.8 & 24.87 & 39.2 & 27.3 &8.1  & 33.29 & 55.6 & 60.7 & 37.5 & 32.1& 20.2 & 21& 27 & 20.9 & 24.6\\
        
        Qwen2.5-VL-7b~\cite{bai2025qwen2} & 4& 33.07 & 28.75& 31.33 & 33.66 & 21.99 & 14.97 & 42.88 & 37.73 & 23.83 & 23.64 & 22.97&33.33 & 28.75 & 6.83 & 40.27&58.24 & 51.49 & 44.75 & 50.00 & 32.13 & 33.87 & 32.85 & 27.15 & 31.93\\

        LLaVA-v1.5-7b~\cite{liu2024improved} & 18& 23.65 & 10.85& 5.17 & 12.53 & 17.37 & 11.34 & 7.25 & 5.26 & 18.73 & 9.12 & 26.50&24.43 & 26.75 & 28.31 & 34.09&51.18 & 52.38 & 34.25 & 24.18 & 26.87 & 34.68 & 29.94 & 22.52 & 30.81\\

        LLaVA-v1.6-7b~\cite{liu2024llavanext} & 19& 13.21 & 8.53& 12.14 & 0.00 & 20.35 & 0.27 & 10.76 & 0.41 & 24.27 & 0.00 & 4.79&6.62 & 7.75 & 0.00 &  20.18&51.76 & 7.74 & 6.25 & 32.14 & 6.37 & 39.52 & 10.47 & 21.52 & 5.88\\
 
        SpaceMantis-13B~\cite{remyxai_spacemantis13b_2025} & 14 & 28.93  & 23.56& 35.2 & 29.1 & 18.1 & 13.3 & 21.4 & 23.1 & 24.9 & 23.4 & 23.27 &31.8 & 31.8 & 6.2 & 35.60 & 56.5 & 55.4 & 41.8 & 31 & 36.3 & 25.3 & 25.6 & 25.5& 23 \\
        
        SpaceQwen2.5-VL-3B~\cite{spaceqwen2025} & 17 & 24.24  & 14.46  & 23.7& 25.8 & 18.9 & 19.1 &4.7 & 3.3 & 13.1 & 7.1 & 24.00 & 34.4 & 29.5 & 8.1 & 33.01 & 58.5 & 58.6 & 35.5 & 23.9 & 24.1 & 22.6 & 27.9 & 19.9 & 26.1 \\

        Spatial-MLLM-7B~\cite{wu2025spatialmllmboostingmllmcapabilities} & 6 & 32.15 & 29.88 & 31.9 & 22.9 & 22.8 & 16.4 & 35.9 & 38.7 & 35.5 & 34.9 & 20.30 & 34.1 & 26.8 & 0 & 38.13 & 54.7 & 50.9 & 39 & 34.6 & 24.7 & 38.7 & 41.3 & 28.8 & 30.5 \\

        VLM3R-7B~\cite{fan2025vlm} & 2 & \cellcolor{lightorange}43.21 & \cellcolor{lightorange}39.78 & 47.8 & 45.6 & 40.1 & 20.6 & 42.2 & 44.3 & 40.1 & 37.5 & 28.43 & 42 & 30 & 13.3 & \cellcolor{lightorange}51.18 & 55.9 & 59.2 & 58.8 & 53 & 54.6 & 47.3 & 50.6 & 30.5 & 50.7 \\
        
        \hline
        \model-SR & 1 &\cellcolor{lightred} 54.87 & \cellcolor{lightred}59.99 & 80.27 & 73.75 & 21.42 & 18.87 & 78.44 & 75.17 & 68.44 & 63.55 & \cellcolor{lightred}36.27 & 27 & 28.2 & 53.3 & \cellcolor{lightred}56.51 & 53.5 & 49.1 & 76.8 & 50 & 68.7 & 50.5 & 52.6 & 44.4 & 63 \\
    \end{tabular}
}
     \caption{\textbf{Performance of different models on SPAR-Bench} The highest, second-highest, and third-highest scores in each category are highlighted with \colorbox{lightred}{light red}, \colorbox{lightorange}{light orange},  and \colorbox{lightyellow} {light yellow}, respectively.  SPAR-Bench (\textit{tiny}) refers to a subset of the full benchmark, where 50 questions are sampled per task. Our model, \model-SR, demonstrate the best performance consistently across all tasks. Notably, it surpasses human performance in low category.}
    \label{tab:sparbench_res_detail}
    \vspace{-5mm}
\end{table*}

\end{document}